\documentclass[10pt,twocolumn,letterpaper]{article}

\usepackage{cvpr}              

\usepackage[dvipsnames]{xcolor}

\usepackage{times}
\usepackage{epsfig}
\usepackage{graphicx}
\usepackage{amsmath}
\usepackage{amssymb}
\usepackage{multirow}
\usepackage{booktabs}
\definecolor{cvprblue}{rgb}{0.21,0.49,0.74}
\usepackage[pagebackref,breaklinks,colorlinks,citecolor=cvprblue]{hyperref}

\def\paperID{51} 
\def\confName{3DV\xspace}
\def\confYear{2024\xspace}

\title{HumanReg: Self-supervised Non-rigid Registration of Human Point Cloud}

\author{Yifan Chen\qquad Zhiyu Pan\qquad Zhicheng Zhong\qquad Wenxuan Guo\qquad Jianjiang Feng\thanks{Corresponding author}\qquad Jie Zhou\\
Department of Automation, Tsinghua University, Beijing, China\\
{\tt\small \{chenyf21, pzy20, zhongzc18, gwx22\}@mails.tsinghua.edu.cn  \qquad \{jfeng,jzhou\}@tsinghua.edu.cn}}

\begin{document}
\maketitle

\begin{abstract}
    In this paper, we present a novel registration framework, HumanReg, that learns a non-rigid transformation between two human point clouds end-to-end. We introduce body prior into the registration process to efficiently handle this type of point cloud. Unlike most exsisting supervised registration techniques that require expensive point-wise flow annotations, HumanReg can be trained in a self-supervised manner benefiting from a set of novel loss functions. To make our model better converge on real-world data, we also propose a pretraining strategy, and a synthetic dataset (HumanSyn4D) consists of dynamic, sparse human point clouds and their auto-generated ground truth annotations. Our experiments shows that HumanReg achieves state-of-the-art performance on CAPE-512 dataset and gains a qualitative result on another more challenging real-world dataset. Furthermore, our ablation studies demonstrate the effectiveness of our synthetic dataset and novel loss functions. Our code and synthetic dataset is available at \url{https://github.com/chenyifanthu/HumanReg}.
    
    \end{abstract}
\section{Introduction}
\begin{figure}
   \centering
   \includegraphics[width=0.98\linewidth]{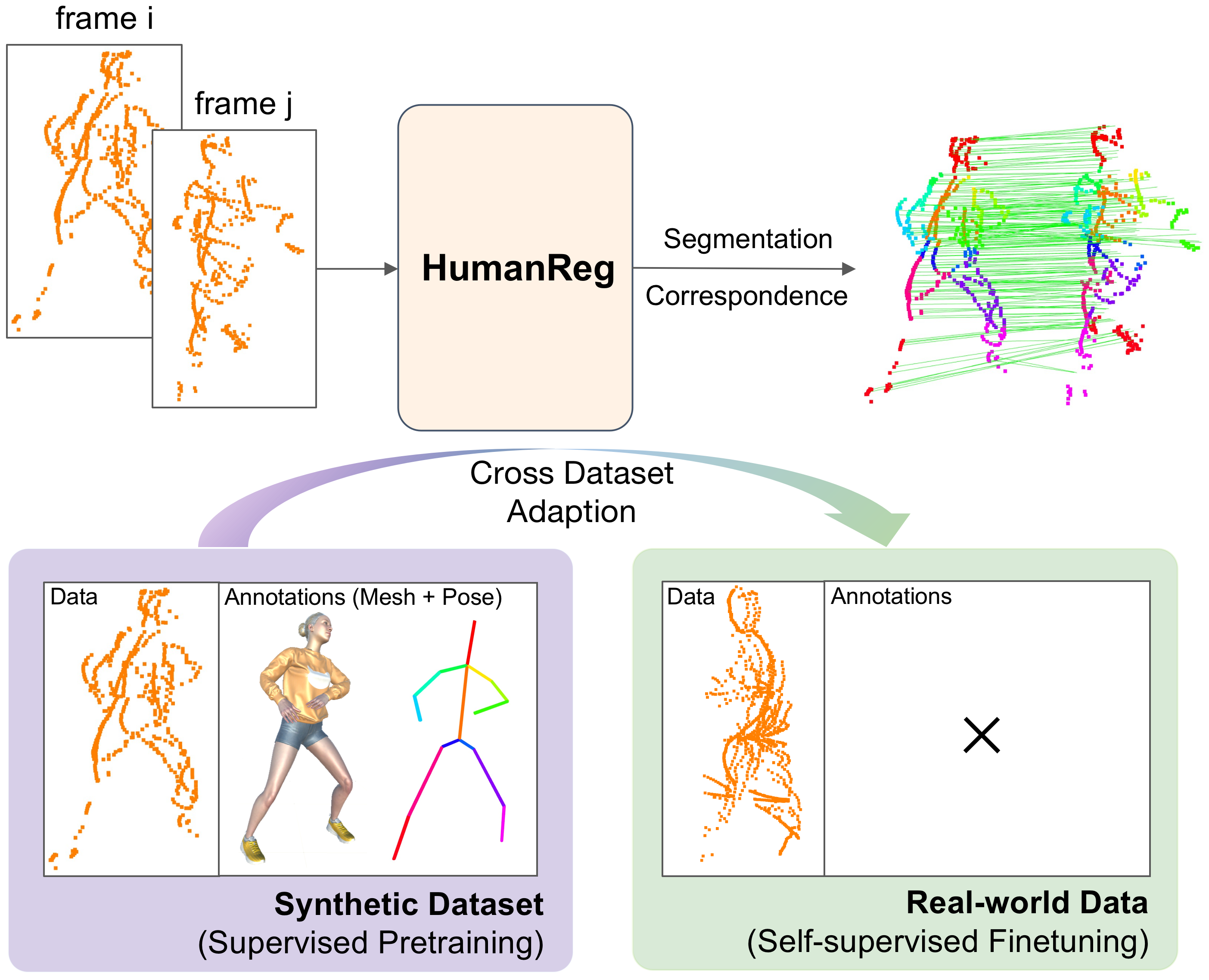}
   \caption{\textbf{HumanReg overview.} The proposed \emph{HumanReg} framework takes a pair of human point clouds as input, simultaneously estimates the body-part segmentation for each point cloud and the scene flow between them. \emph{HumanReg} can be pretrained on our synthetic dataset using ground-truth annotations, then adapted to unlabeled real-world data with our proposed self-supervised loss.}
\label{fig:overview}
\end{figure}

Point cloud is a crucial data format in the fields of robotics and autonomous driving, where robots need to capture and analyze data from the environment dynamically. In indoor scenes, depth cameras \cite{dou2013scanning}, dense IR cameras \cite{collet2015high} or a set of multi-view RGB cameras \cite{joo2015panoptic} are commonly used to record dynamic 3D data of the scene or objects in it. As to the outdoors, considering the need of large covering, existing 3D imaging systems \cite{geiger2013vision, caesar2020nuscenes, sun2020scalability, patil2019h3d, geyer2020a2d2, choi2018kaist, chang2019argoverse, cong2022stcrowd} often use LiDAR to achieve real-time scanning of the surroundings. This inevitably brings two problems. First, in each individual frame, the point cloud of each object is sparse due to the large distance from scanning device. In addition, such point clouds often suffer from occlusion, ambiguity, and noise. Both issues affect downstream tasks performance such as 3D reconstruction \cite{izadi2011kinectfusion, lin2018learning}, human pose estimation \cite{li2022lidarcap, zhou2020learning, bekhtaoui2020view}, and object recognition \cite{maturana2015voxnet}.

A simple idea to densify dynamic object point clouds is to align the scanned frames at different times. For rigid objects like cars in the scene, the problem can be formulated as estimating the rigid transformation matrix between any two frames. Many recent works have achieved good results on this problem using traditional \cite{besl1992method, zhou2016fast, yang2015go, rusu2009fast} or 
learning-based methods \cite{aoki2019pointnetlk, choy2020deep, pais20203dregnet, xu2021omnet, yew2020rpm}. As for deformable objects like human bodies, a feasible method is to estimate the scene flow \cite{vedula1999three}, which describes the 3D motion for each point. Most of the existing learning-based methods aims at minimizing the flow loss under the ground truth supervision of scene flow \cite{behl2019pointflownet, gu2019hplflownet, puy2020flot, ushani2018feature, huang2021multibodysync} or point correspondences \cite{li2022lepard, huang2020consistent, marin2020correspondence, donati2020deep}. However, obtaining scene flow annotations for real-world data is quite expensive. Although some works \cite{jacobson2014skinning,loper2015smpl} try to fit a template mesh to human point cloud and calculate the scene flow, such methods will introduce non-negligible error on sparse cases \cite{li2022lidarcap}.

To resolve such problems, we propose HumanReg (Fig. \ref{fig:overview}), a non-rigid registration framework designed to register human point clouds captured from the same person at different times. Different from previous works that directly estimate the correspondence between two point clouds, we introduce a body-part segmentation head into our framework. This provides latent pose information for point-wise feature learning and benefits the registration process.

Registering rigid objects is usually easier than non-rigid ones because we only need to estimate a 6DoF transformation matrix. The transformation of human body is a special case, which can be regard as a combination of several rigid parts. This is a valid assumption because each skeleton of human body is not deformable. Therefore, we can use the body-part segmentation results to guide the registration process. We formulate such rigid constraints as a novel self-supervised loss function, which can be used to train our model without the need of expensive ground-truth labels.

To give the model a better initialization and convergence on real-world data, we propose a multi-person and multi-view synthetic dataset, HumanSyn4D, to simulate human point clouds scanned by mechanical LiDARs in a large outdoor scene. Since the ground-truth correspondence and pose of each person can be generated automatically from the known template meshes, we use this dataset to pretrain both heads of our framework. Then we adapt the model to real-world data using the proposed self-supervised loss. Our experiments show that the combination of our framework, synthetic dataset, pretraining and finetuning strategy can achieve remarkable results on diffent types of real-world data. In summary, our contributions are:
\begin{itemize}
\item We propose an end-to-end human point cloud registration framework, HumanReg, that predicts the scene flow between raw human point clouds. We introduce body-part segmentation head into our framework to enhance extracted features.
\item We formulate the non-rigid human registration problem as a part-rigid registration problem and design a novel self-supervised loss to train our model without the need of expensive ground-truth labels.
\item We propose a multi-person synthetic dataset, HumanSyn4D, to make our model better converge on real-world data. 
\end{itemize}
\section{Related Work}
\subsection{Non-rigid Point Cloud Registration}
\noindent{\textbf{Correspondence-based Method}}. Finding accurate point correspondences between point clouds is a useful solution in both rigid and non-rigid registration task. At scene level, DynamicFusion \cite{newcombe2015dynamicfusion} finds correspondences by matching the depth map of adjacent frame based on a coarse warping field. VolumeDeform \cite{innmann2016volumedeform} computes SIFT \cite{lowe2004distinctive} matches of input frames to improve tracking quality. Schmidt et al. \cite{schmidt2016self} also use an image-based method to extract features and reconstruct scenes. At object level, traditional methods \cite{amberg2007optimal, myronenko2010point, yao2020quasi} aim at minimizing certain type of optimization functions. Inspired by some rigid registration frameworks \cite{ao2021spinnet, deng2018ppfnet, huang2021predator} using learned local or global descriptors to extract point features, 3DCODED \cite{groueix20183d} learns a global vector to transform the template into input surface. Lepard \cite{li2022lepard} enhances extracted point features with a self-attention and cross-attention module. 

\noindent{\textbf{Scene Flow Estimation}}. Scene flow \cite{vedula1999three} directly describes the 3D transition between two point clouds. A few methods \cite{behl2019pointflownet,huang2022dynamic} split the scene point cloud into a static background and rigid objects with different motions to obtain scene flow. FlowNet3D \cite{liu2019flownet3d} applies a flow embedding layer and a set upconv layer to estimate flow end-to-end. FLOT \cite{puy2020flot} and PointPWC-Net \cite{wu2019pointpwc} use network to estimate the transportation distance and cost volume between two point clouds. PointPWC-Net also proposes a self-supervised loss to train the model. Some recent techniques \cite{donati2020deep, melzi2019zoomout, huang2022multiway} handle this problem by aligning shapes via functional map \cite{ovsjanikov2012functional}.

\subsection{Non-rigid 3D Datasets}
Unlike rigid dataset \cite{wu20153d, chang2015shapenet, zeng20173dmatch} whose point clouds are directly sampled from the surface of static object or scene, non-rigid dataset contains deforming objects and their sequences of motion. 

\noindent{\textbf{Real-world Dataset}}. Most of the real-world collected datasets focus on reconstructing the surface of human body \cite{anguelov2005scape, vlasic2008articulated, guo2015robust, ye2012performance, innmann2016volumedeform, bozic2020deepdeform, zheng2019deephuman}. Their original data is mainly collected from RGB-D cameras, and they are relatively small, and the collection device has to be placed close enough to the human body to get a dense scan. In auto-driving field, KITTI \cite{geiger2012we, geiger2013vision} collects a wide range of outdoor 3D scans with LiDARs fixed to a moving car. Based on KITTI, Menze et al. \cite{menze2018object, menze2015joint} estimate 2D optical flow and project it to 3D point cloud to get sparse scene flow.

\noindent{\textbf{Synthetic Dataset}}. Although LiDAR scans can cover a large area, labeling point-wise scene flow annotations is always a challenging and error-prone task. Synthetic method has been used to solve this dilemma in recent work \cite{butler2012naturalistic, mayer2016large, li20214dcomplete, li2022lepard}. Among them, FlyingThing3D \cite{mayer2016large} uses Blender to generate random 3D trajectories for everyday objects. DeformingThings4D \cite{li20214dcomplete} introduces a large synthetic dataset, covering a wide variety of deforming things from humanoids to animal species. As for human body, SMPL \cite{loper2015smpl} uses a skinned vertex-based model to generate naked body mesh of different body shapes and poses. Ma et al. \cite{ma2020learning} proposes a framework to represent clothed human body, and uses a high-resolution body scanner to obtain dense scan sequences. However, the form of their point clouds are quite different from LiDAR data collected in a real-world large scene.
\section{Method}

\subsection{Problem Definition}
Given a pair of human point clouds $\mathbf{P}\in \mathbb{R}^{n\times 3}$ and $\mathbf{Q}\in\mathbb{R}^{m\times 3}$, where $n, m$ are the number of points, our goal is to find a warp function $\mathcal{W}: \mathbb{R}^{n\times 3} \mapsto \mathbb{R}^{n\times 3}$ that aligns $\mathbf{P}$ to $\mathbf{Q}$. In this work, we solve this problem by estimating per-point 3D flow $\mathbf{F}\in \mathbb{R}^{n\times 3}$, and the warp functions can be defined as $\mathcal{W}(\mathbf{P}) := \mathbf{P} + \mathbf{F}$. 

Given a human point cloud $\mathbf{P}$ and its corresponding ground truth pose, we can assign a label $l_i$ for each point $\mathbf{p}_i \in \mathbf{P}$, represented the body part it belongs to
\begin{equation}
   \label{eq:problem-definition}
   l_i = \mathop{\arg\min}\limits_{k\in 1,\ldots,K}  d(\mathbf{p}_i, B_k),
\end{equation}
where $B_k$ represents the $k$-th segment skeleton of the human body (from the given 3D joint locations and their topological connection method), and $d(\cdot, \cdot)$ calculate the distance from a point to line segment in $\mathbb{R}^3$ space. In this work, we use 15 joint points to represent body skeleton, and the detailed definition is provided in Suppl.

\subsection{Architecture of HumanReg}
\label{sec:model-architecture}
\begin{figure*}
   \centering
   \includegraphics[width=17.3cm]{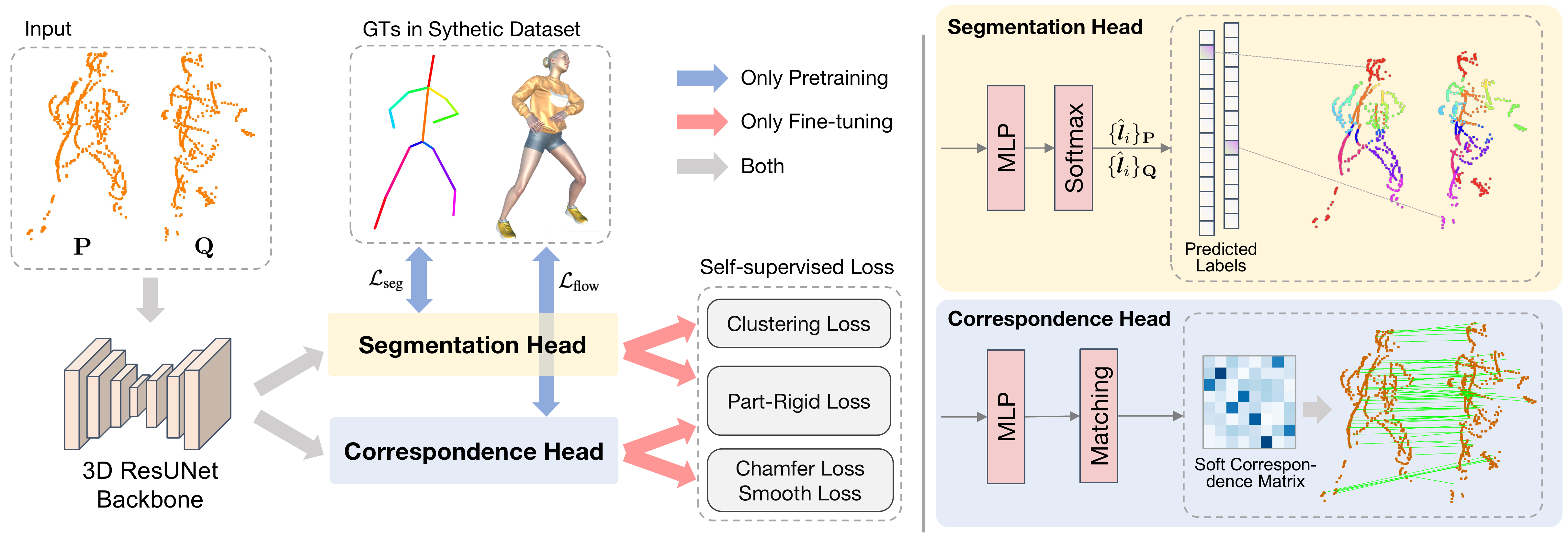}
\caption{\textbf{Training pipeline of our proposed method.} Given the input human point clouds $\mathbf{P}$ and $\mathbf{Q}$, the 3D ResUNet backbone extracts per-point features, which are then processed by a segmentation head and a correspondence head (Sec. \ref{sec:model-architecture}). The two heads simultaneously output body-part segmentation of each point cloud and the soft correspondence between them. Our model is firstly pretrained on synthetic dataset with ground-truth labels and flow (Sec. \ref{sec:supervised-pretraining}). Then, a set of self-supervised loss functions (Sec. \ref{sec:self-supervised-loss}) are applied based on the estimation of both heads when finetuning on real-world data.}
   \label{fig:framework}
\end{figure*} 

Fig. \ref{fig:framework} shows the overview of our proposed method.

\noindent \textbf{Backbone.} We utilize the ResUNet backbone \cite{choy2020deep} to extract point descriptors of input point clouds. The backbone is implemented with MinkovskiEngine \cite{choy20194d}, which defines standard neural network layers like convolutional and deconvolutional layer on 3D data, and uses sparse tensor to speedup inference and minimize memory footprint. In human registration task, we fix the size of each sparse voxel at 0.01m.

\noindent \textbf{Segmentation Head.} The extracted descriptors, denoted as $\mathbf{D}_i$, are passed through a body-part segmentation head utilized by an MLP and a softmax layer, where the predicted label for each point can be defined as
\begin{equation}
   \hat{l_i} = \text{Softmax}(\text{MLP}(\mathbf{D}_i)).
\end{equation}

In segmentation head, we introduce human body prior to the custom registration model. This has two advantages: 1) The body-part information can enhance the features extracted by the backbone and reduce mismatch in the correspondence head. 2) Segmentation and flow estimation will be combined to compute our self-supervised loss (Sec. \ref{sec:self-supervised-loss}).

\noindent \textbf{Correspondence Head.} The extracted descriptors $\mathbf{D}_i$ are simultaneously passed through another head to estimate flow. The descriptors are first updated by an MLP layer: $\mathbf{D}_i \leftarrow \text{MLP}(\mathbf{D}_i)$. We use soft correspondence to describe the relationship between the inputs. Given the input updated descriptor $\mathbf{D}^\mathbf{P},\mathbf{D}^\mathbf{Q}$, the soft correspondence matrix $\mathbf{C}\in\mathbb{R}^{n\times m}$ can be computed as 
\begin{equation}
   \mathbf{C}_{ij} = - \frac{1}{t} \left\Vert \mathbf{D}^\mathbf{P}_i - \mathbf{D}^\mathbf{Q}_j \right\Vert_2,
\end{equation}
\begin{equation}
    \mathbf{C}_{i:} \leftarrow \text{Softmax}(\mathbf{C}_{i:}).
\end{equation}

Here, a trainable parameter $t$ is used to control the distance threshold in training process. We set its initial value to 0.1 and minimum value to 0.02. The softmax function is applied to each row in $\mathbf{C}$ to make $\mathbf{C}$ row-stochastic. We can obtain flow estimation by 
\begin{equation}
   \mathbf{F} = \mathbf{P}^w - \mathbf{P} = \mathbf{CQ} - \mathbf{P},
\end{equation}
where $\mathbf{P}^w$ is the warped point cloud of $\mathbf{P}$.

\subsection{Supervised Pretraining on Synthetic Data}
\label{sec:supervised-pretraining}
It's easy to aquire ground-truth labels and flow of the synthetic dataset (Sec. \ref{sec:humansyn4d}) where the template mesh of each avatar is known. Therefore, we use supervised learning to pretrain our model on synthetic data.

In the segmentation head, the human pose is used to generate ground-truth labels according to Eq. \ref{eq:problem-definition}. Then, we use CrossEntropy criterion $\text{CE}(\cdot,\cdot)$ to calculate the supervised loss
\begin{equation}
   \mathcal{L}_{\text{seg}} = \frac{1}{n}\sum_{i=1,\ldots,n}\text{CE}(\hat{l_i}, l_i^\text{gt}).
\end{equation}

In the correspondence head, we directly measure the flow loss as the Frobenius norm between ground-truth flow annotation $\mathbf{F}^\text{gt}$ and the estimated flow $\mathbf{F}$:
\begin{equation}
   \mathcal{L}_{\text{flow}} = \frac{1}{n}\left\Vert \mathbf{F}_i - \mathbf{F}^\text{gt}_i \right\Vert_F^2,
\end{equation}

Finally, we use weighted parameters $\alpha_1, \alpha_2$ to balance the total pretrain loss 
\begin{equation}
   \mathcal{L}_{\text{pt}} = \alpha_1\mathcal{L}_\text{seg} + \alpha_2\mathcal{L}_\text{flow},
\end{equation}

\subsection{Self-supervised Finetuning on Real-world Data}
\label{sec:self-supervised-loss}
It is unrealistic to manually label scene flow directly on real-world collected point clouds. In this section, we propose a self-supervised objective function designed specifically for HumanReg. It consists of four parts: \emph{Chamfer Loss}, \emph{Smoothness Loss}, \emph{Clustering Loss}, and \emph{Part-Rigid Loss}. 

\noindent{\textbf{Chamfer Loss.} } Chamfer loss encourages source point cloud warped close to the target.

\begin{equation}
\begin{aligned}
   \mathcal{L}_{\text{chamfer}} = & \frac{1}{n} \sum_{\mathbf{p}_i^w \in \mathbf{P}^w} \min_{\mathbf{q}_j \in \mathbf{Q}}{\left\Vert \mathbf{p}_i^w - \mathbf{q}_j \right\Vert_2^2} + \\
   & \frac{1}{m} \sum_{\mathbf{q}_j \in \mathbf{Q}} \min_{\mathbf{p}_i^w \in \mathbf{P}^w}{\left\Vert \mathbf{p}_i^w - \mathbf{q}_j \right\Vert_2^2}.
\end{aligned}
\end{equation}

\noindent{\textbf{Smoothness Loss.}} Inspired by \cite{wu2019pointpwc}, smoothness loss enforces local spatial smoothness, which means close points in space should have similar flows.
\begin{equation}
   \label{eq:smoothness-loss}
   \mathcal{L}_{\text{smooth}} = \frac{1}{n}\sum_{\mathbf{p}_i \in \mathbf{P}} \frac{1}{|\mathcal{N}_{\mathbf{P}}(\mathbf{p}_i)|} \sum_{\mathbf{p}_j \in \mathcal{N}_{\mathbf{P}}(\mathbf{p}_i)} \left\Vert \mathbf{F}_i - \mathbf{F}_j \right\Vert_2^2,
\end{equation}
where $\mathcal{N}_{\mathbf{P}}(\mathbf{p}_i)$ is the point set of $k$ nearest neighbors of $\mathbf{p}_i$.

\noindent{\textbf{Clustering Loss.}} Similar to smoothness loss, clustering loss assumes that points belonging to the same body-part should cluster together in space. It can be written as
\begin{equation}
   \label{eq:clustering-loss}
   \mathcal{L}_{\text{cluster}} = \frac{1}{n}\sum_{\mathbf{p}_i \in \mathbf{P}} \frac{1}{|\mathcal{N}_{\mathbf{P}}(\mathbf{p}_i)|} \sum_{\mathbf{p}_j \in \mathcal{N}_{\mathbf{P}}(\mathbf{p}_i)} \text{CE}\left(\hat{l_i}, \hat{l_j}\right).
\end{equation}

\begin{figure}
   \centering
   \includegraphics[width=7cm]{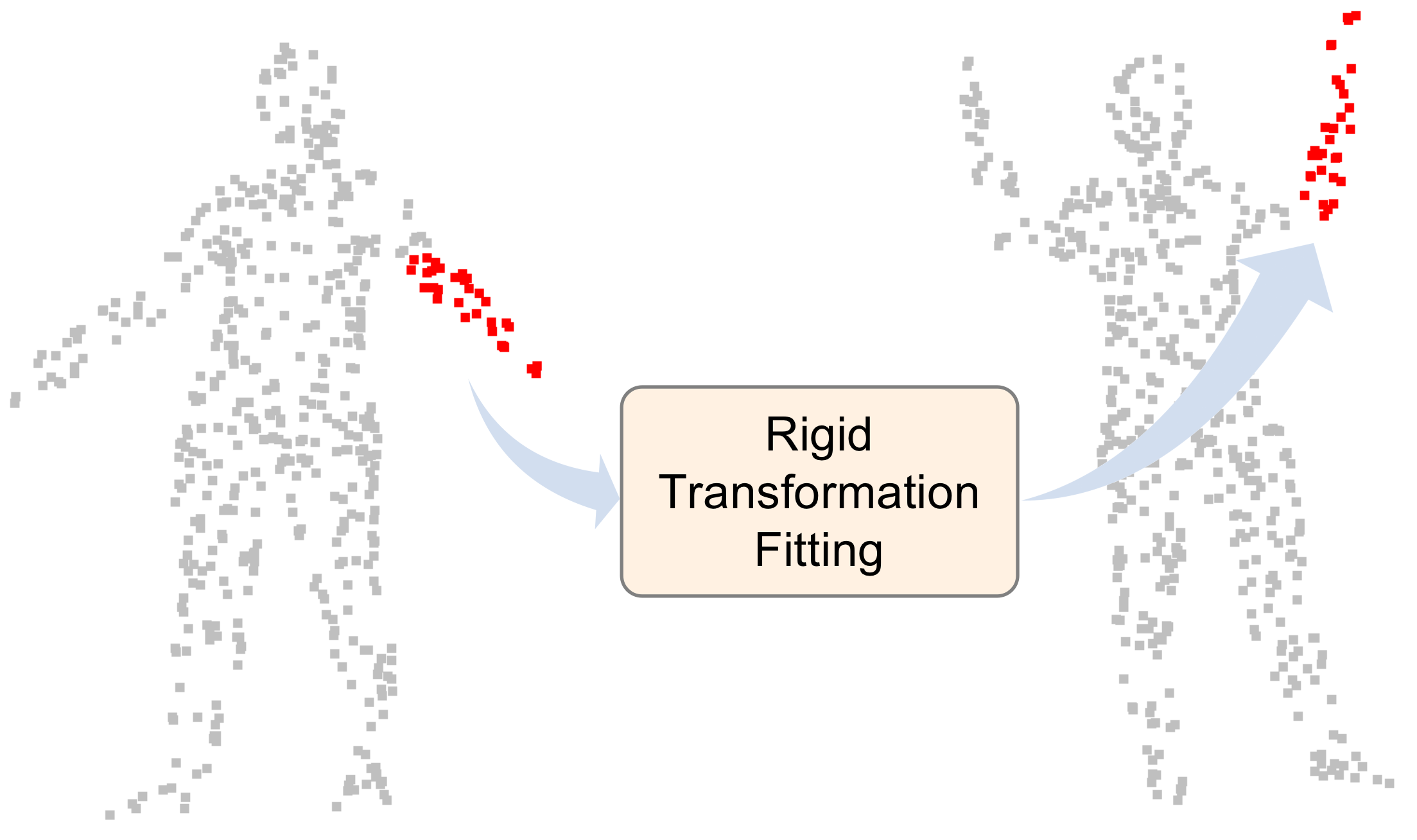}
   \caption{\textbf{Rigid fitting for body part.} We assume that the warp field of each body part is close to a rigid transformation. This assumption is used to design our part-rigid loss and refine flow during test time.}
   \label{fig:rigid-assumption}
\end{figure}

\noindent{\textbf{Part-Rigid Loss.}} Solving the non-rigid registration problem is always more difficult than the rigid one. The rigid transformation is a 6-DoF matrix, while non-rigid warp field is composed of per-point 3D flow. However, the human body is somewhat between rigid and non-rigid. In SMPL model \cite{loper2015smpl}, a local part has different mesh deformations as the pose changes. But for point clouds scanned in a large scene, the slight inconsistency is ignorable compared to the sensor noise. In this way, we assume that the warp field of each body part can be approximated by a rigid transformation.

In our model, we design a segmentation head. It not only helps to enhance extracted feature during pretraining, but also divides the original point cloud into several body parts. Thus, utilizing the output of both head in our model, we formulate the above assumption as a part-rigid loss. As shown in Fig. \ref{fig:rigid-assumption}, for each body part predicted in segmentation head, we first estimate a rigid transformation for its warp field
\begin{equation}
   \label{eq:fitting}
   \mathbf{T}_k^* = \mathop{\arg\min}\limits_{\mathbf{T}_k} \left\Vert \mathbf{T}_k \circ \mathbf{P}_k - (\mathbf{P}_k + \mathbf{F}_k) \right\Vert_2,
\end{equation}

where $\mathbf{P}_k, \mathbf{F}_k \in \mathbb{R}^{n_k \times 3}$ represents the point set of the $k$-th body part and the estimated flow output by correspondence head. $\mathbf{T}_k^*$ can be decomposed into a rotation matrix $\mathbf{R}_k \in SE(3)$ and a translation vector $\mathbf{t}_k \in \mathbb{R}^3$. Our part-rigid loss describes the fitting error between the rigid transformation and the estimated scene flow of each body part
\begin{equation}
   \mathcal{L}_{\text{rigid}} = \frac{1}{n}\sum_{k=1}^K \sum_{\mathbf{p}_i \in \mathbf{P}_k} \left\Vert (\mathbf{R}_k - \mathbf{I}) \cdot \mathbf{p}_i + \mathbf{t}_k - \mathbf{f}_i \right\Vert_2^2.
\end{equation}

The total self-supervised loss is the weighted sum of the four type of losses
\begin{equation}
   \mathcal{L}_{\text{total}} = \sum \beta_{\text{type}} \mathcal{L}_{\text{type}},
\end{equation}
where type $\in \{\text{chamfer, smooth, cluster, rigid}\}$. 

During test time, we use the same assumption in Fig. \ref{fig:rigid-assumption} to refine flow estimation. After computing the rigid transformation $\mathbf{R}_k, \mathbf{t}_k$ using Eq. \ref{eq:fitting}, the final flow output for each body part is
\begin{equation}
   \tilde{\mathbf{F}_k} = \mathbf{R}_k \cdot \mathbf{P}_k + \mathbf{t}_k - \mathbf{P}_k.
\end{equation}

\begin{figure*}
   \centering
   \includegraphics[width=0.97\linewidth]{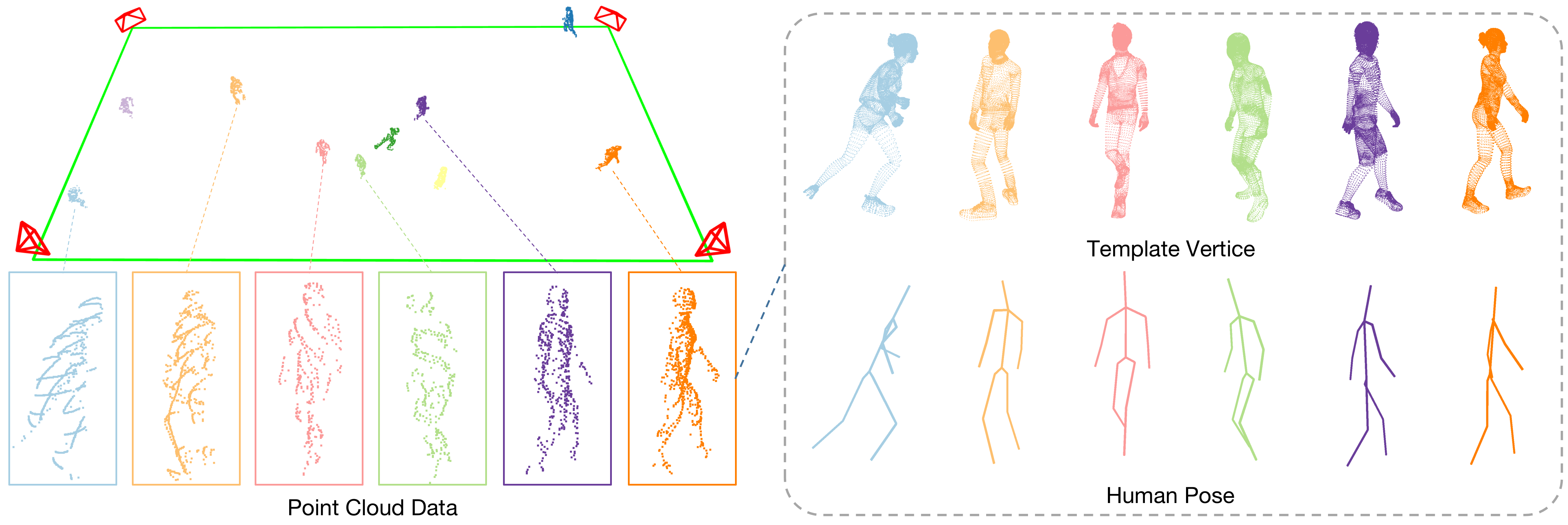}
   \caption{\textbf{A snapshot of HumanSyn4D.} Top Left: Top view of the synthetic scene, \includegraphics[height=0.3cm]{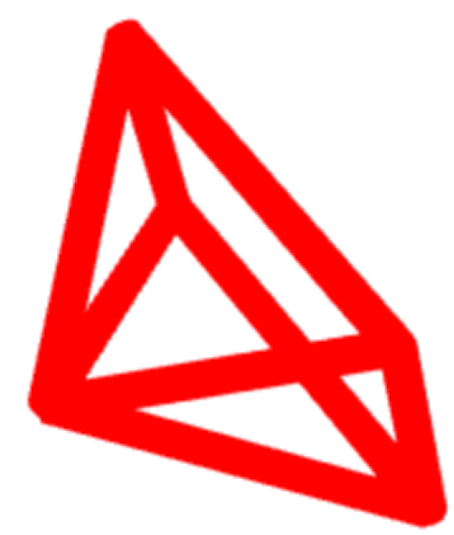} represents simulated LiDAR and green lines are the boundaries of the field. Bottom Left: Scanned human point clouds in one frame. Top Right: Ground-truth template mesh vertices. Bottom Right: Ground-truth human pose of each person.}
   \label{fig:humansyn4d}
\end{figure*}

\section{HumanSyn4D}
\label{sec:humansyn4d}

Training our network in a supervised manner requires a sufficient amount of human data with pose and flow ground truth at point level. To provide such data, we propose HumanSyn4D, a multi-person multi-view synthetic dataset consisting of sparse point clouds, avatar's mesh vertices in any timestamp, and 3D human pose labels. A snapshot of HumanSyn4D is shown in Fig. \ref{fig:humansyn4d}.

We develop our synthetic system on Unity platform \cite{juliani2018unity} due to its flexibility and productivity. Specifically, we download ten different human 3D models from Adobe Mixamo\footnote{\url{https://www.mixamo.com}} and initially place them randomly in a $30 \text{m} \times 15 \text{m}$ scene. We use action files to drive the deformation of the human mesh and update each person's position in the scene.

To collect human point clouds, we place four simulated LiDARs at the four corners of the scene. The laser beam is emitted from the center of the LiDAR at a certain angle, falls on a human mesh surface and returns its distance. This acquisition method effectively simulates occlusion in the real world. We use non-repetitive sampling of the Livox Mid-40 LiDAR to emit laser beam, which can be formulated as 
\begin{equation}
   r = r_0 \cos(\omega t + \theta_0),
\end{equation}
where $\omega$ is the angular velocity of the wedge rotation in LiDAR, $r_0$ is the maximum scanning radius in pixel and $\theta_0$ is a random initial angle. This equation is defined in polar coordinates on a virtual imaging plane. We eventually convert it to Cartesian coordinates and merge the points of each person.

\section{Experiments}
\subsection{Dataset and Settings}
\begin{figure}
   \centering
   \includegraphics[width=0.9\linewidth]{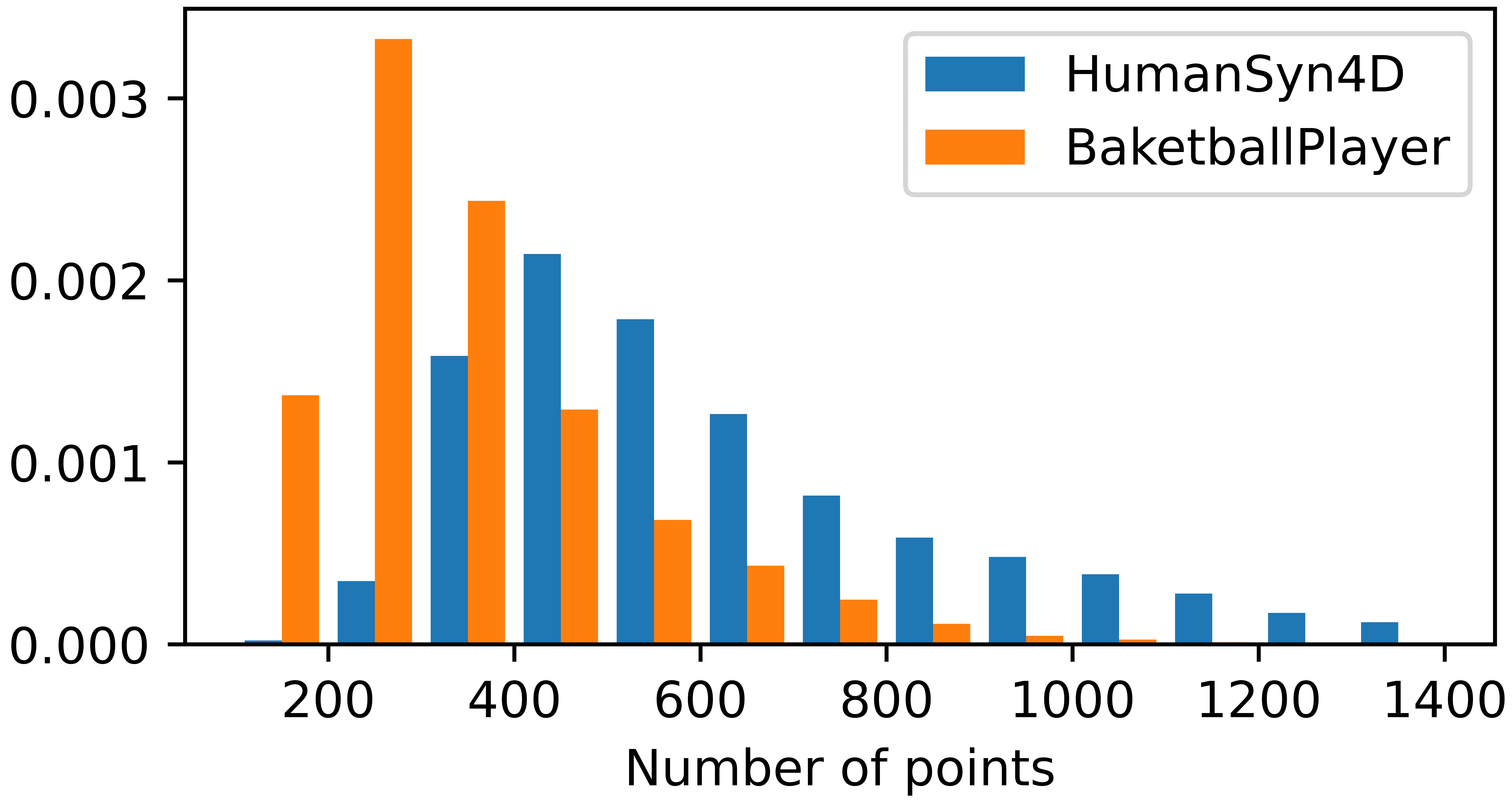}
   \caption{Histogram of numbers of points in HumanSyn4D and BasketballPlayer.}
   \label{fig:npoints}
\end{figure}

\noindent \textbf{Datasets.} Our experiments are conducted on two kind of human point cloud dataset, For all the datasets we keep only the points from the foreground human body.
\begin{itemize}
   \item \textbf{CAPE-512.} CAPE dataset \cite{ma2020learning} contains 3D human body point clouds scanned with a high-resolution body scanner. Huang et al. \cite{huang2022multiway} sample from its raw scans and obtain the ground-truth flow from the fitted template mesh to compose the MPC-CAPE dataset. However, each body scan in MPC-CAPE has 8192 points, which is much denser than the point cloud of outdoor scans. Therefore, we randomly sample 512 points (1/16 of the original resolution) in each scan to form the CAPE-512 dataset. We utilize the mesh template estimated from dense point clouds to obtain the ground-truth flow. CAPE-512 is used for quantitative comparison with baselines.
   \item \textbf{BasketballPlayer} is a much more challenging real-world dataset collected by ourselves. We use four Livox Mid-100 LiDARs to record a basketball match with ten players. Due to the fast movement and fierce confrontation of the players, there are large noises and occlusions in the data. We first calibrate the external parameters between four LiDARs, then crop the original scan to remove points from the surroundings and use a fitted plane to remove the ground. Points from different people are separated and tracked throughout the match. The density distribution of HumanSyn4D and Basketballplayer is shown in Fig. \ref{fig:npoints}.
\end{itemize}  

The comparison of the two datasets and our synthetic dataset is shown in Table \ref{tab:comparison-dataset}.

\begin{table}
   \caption{\textbf{Comparison of Datasets.} Dataset division in experiment and differences between three datasets.}
   \begin{center}
   \resizebox*{\linewidth}{!}{
   \begin{tabular}{l|cccc}
   \toprule
   Dataset         & \# Train / Val / Test & Label & Real & \begin{tabular}[c]{@{}c@{}}Large\\ Scene\end{tabular} \\ \midrule
   HumanSyn4D      & 15862 / 4532 / 2266    & \checkmark     & -    & \checkmark       \\
   CAPE-512    & 12060 / 3192 / 836     & \checkmark     & \checkmark    & -       \\
   BaketballPlayer & 13916 / 4000 / 2000    & -     & \checkmark    & \checkmark       \\ \bottomrule
   \end{tabular}
   }
   \end{center}
   \label{tab:comparison-dataset}
\end{table}

\noindent \textbf{Baselines and Training Strategy.} In our work, we focus on comparing the performance of different baselines in self-supervised manner. For fairness, we pretrain all methods on HumanSyn4D dataset and compare their unsupervised training performance on CAPE-512 and BasketballPlayer. We compare the following baselines and formulate different training strategies for them.
\begin{itemize}
   \item \textbf{Non-learned:} Coherent Point Drift method (\underline{CPD}) \cite{myronenko2010point}. We directly test it on target datasets.
   \item \textbf{Supervised Only:} \underline{FLOT} \cite{puy2020flot}. We train it on HumanSyn4D and then test it on target datasets directly.
   \item \textbf{Both Supervised and Self-supervised:} PointPWC-Net (\underline{PointPWC}) \cite{wu2019pointpwc} where we modify the number of points in the feature pyramid to accommodate sparse input, \underline{Synorim} \cite{huang2022multiway} where we use the full pipeline, and our method (\underline{HumanReg}). We pretrain these frameworks on HumanSyn4D in a supervised manner first, then finetune them on target datasets with their self-supervised loss.
\end{itemize}

\noindent \textbf{Evaluation Metrics.} We use following metrics to evaluate flow quality: (1) 3D End-Point Error (EPE3D): flow error $|| \mathbf{F}^\text{gt} - \mathbf{\tilde{F}}||_2$ over all points where $\mathbf{\tilde{F}}$ donates the predicted flow. (2) 3D Accuracy Strict (AccS): the percentage of points with EPE3D $<$ 0.05m. We removed the relative error part in \cite{gu2019hplflownet} because we think this will underestimate the error when the human displacement is large. (3) 3D Accuracy Relax (AccR): the percentage of points with EPE3D $<$ 0.1m. (4) Outlier Ratio: the percentage of points with EPE3D $>$ 0.2m.

\noindent \textbf{Parameters Setting.} Following \cite{huang2022multiway, choy2020deep}, we use a 4-layer U-Net as our backbone, and the output feature dimension is 64. $k$ is set to 5 when searching for neighbors of a point in Eq. \ref{eq:smoothness-loss} and \ref{eq:clustering-loss}. In supervised learning, we balance segmentation and flow loss with weights of $\alpha_1 = 0.1, \alpha_2 = 0.9$. In self-supervised learning, we balance Chamfer / Smoothness / Clustering / Part-Rigid losses with the weights 1.0 / 1.0 / 0.1 / 10.0, respectively. We divide CAPE-512 and BasketballPlayer into non-consecutive sequences of length 4 at equal intervals. In the experiment, we register frames 1-3 with frame 4 of each sequence and calculate metrics.

\subsection{Quantitative Results on CAPE-512}
\begin{table*}
   \caption{\textbf{Quantitative comparison on CAPE-512.} We report the mean and standard deviation metrics of all sequences. $\uparrow / \downarrow$ means higher / lower is better. \emph{w/o refine} is the result without refinement step during test time. The best numbers are highlighted in \textbf{boldface}.}
   \begin{center}
   \resizebox{0.9\linewidth}{!}{
   \begin{tabular}{l|cc|ccccc}
   \toprule
   Method &
    \begin{tabular}[c]{@{}c@{}}Supervised\\ Pretraining\end{tabular} &
    \begin{tabular}[c]{@{}c@{}}Self-supervised\\ Fine-tuning\end{tabular} &
    \begin{tabular}[c]{@{}c@{}}EPE3D $\downarrow$\\ (cm)\end{tabular} &
    \begin{tabular}[c]{@{}c@{}}AccS $\uparrow$\\ (\%)\end{tabular} &
    \begin{tabular}[c]{@{}c@{}}AccR $\uparrow$\\ (\%)\end{tabular} &
    \begin{tabular}[c]{@{}c@{}}Outlier $\downarrow$\\ (\%)\end{tabular} &
    \begin{tabular}[c]{@{}c@{}}Time $\downarrow$\\ (s)\end{tabular} \\ \midrule
   CPD \cite{myronenko2010point}               & -           & - & 9.44 $\pm$ 2.90 & 19.3 $\pm$ 10.8 & 68.5 $\pm$ 19.3 & 4.80 $\pm$ 7.18  & 2.28 \\
   FLOT \cite{puy2020flot}              & flow         & -  & 7.36 $\pm$ 4.87 & 59.0 $\pm$ 24.1 & 81.5 $\pm$ 21.0 & 7.72 $\pm$ 12.07 & 0.16 \\
   PointPWC \cite{wu2019pointpwc}         & flow         & \checkmark  & 6.32 $\pm$ 4.02 & 67.0 $\pm$ 20.2 & 85.0 $\pm$ 15.7 & 7.11 $\pm$ 10.53 & \textbf{0.13} \\
   Synorim \cite{huang2022multiway}           & flow         & \checkmark  & 6.62 $\pm$ 3.32 & 55.2 $\pm$ 17.9 & 84.8 $\pm$ 14.1 & 3.88 $\pm$ 6.29  & 0.63 \\
   \textbf{Ours} (w/o refine) & flow + joint & \checkmark  & 4.54 $\pm$ 1.55 & 66.7 $\pm$ 10.1 & 95.7 $\pm$ 6.2  & 0.51 $\pm$ 3.36  & 0.28 \\
   \textbf{Ours}              & flow + joint & \checkmark  & \textbf{3.22 $\pm$ 1.73} & \textbf{85.6 $\pm$ 11.4} & \textbf{97.4 $\pm$ 6.3}  & \textbf{0.46 $\pm$ 3.65}  & 0.32 \\ \bottomrule
   \end{tabular}
   }
   \end{center}
   \label{tab:cape512-result}
   \end{table*}

\begin{figure*}
   \centering
   \includegraphics[width=0.9\linewidth]{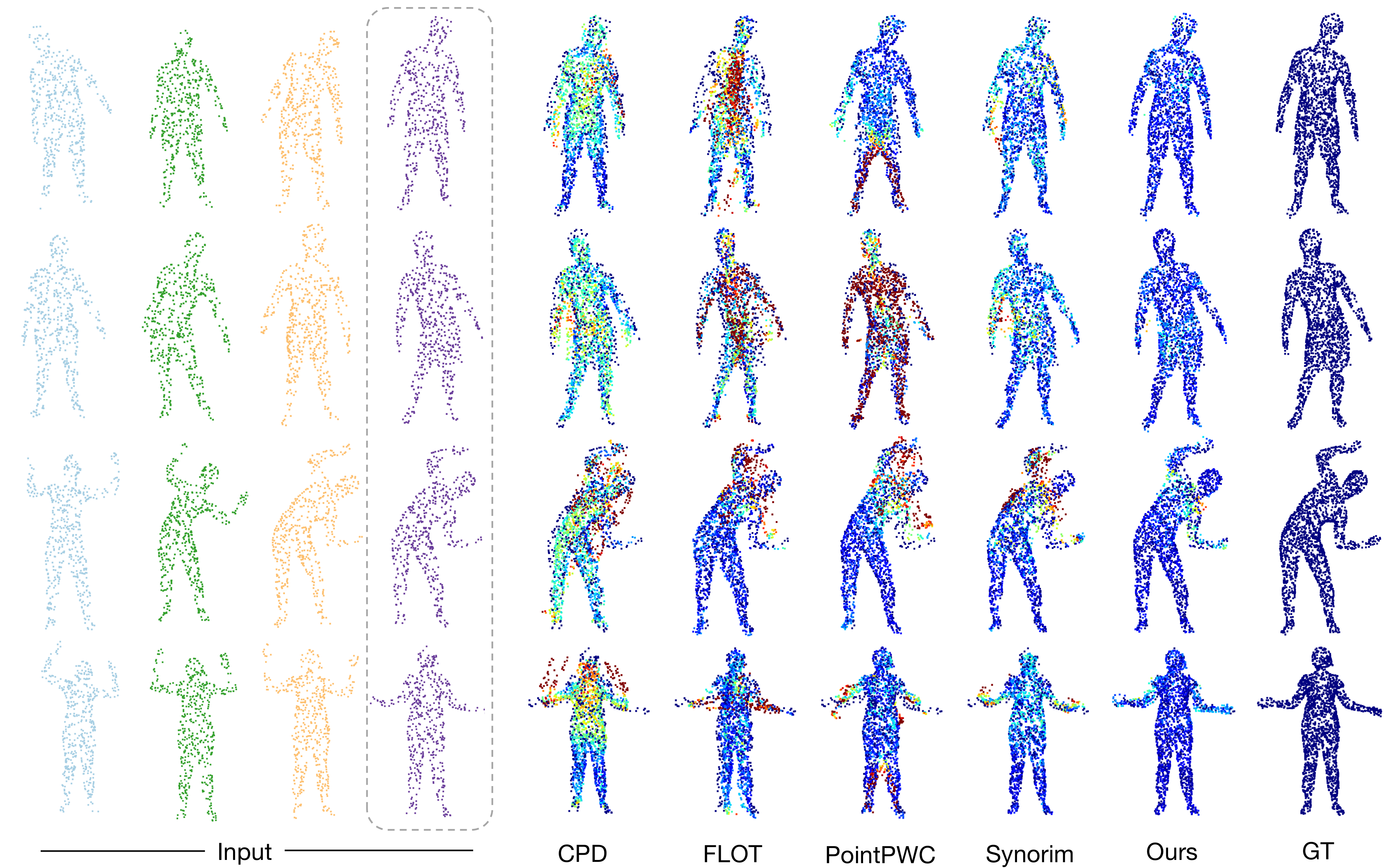}
   \caption{\textbf{Visual comparison results on CAPE-512.} Left: Input 4 frames of sparse point clouds distinguished by 4 different colors. The dashed rectangle is the reference frame to which other frames are aligned. Right: Registered point clouds and per-point L2 error are shown with colormap (0cm \includegraphics[height=0.2cm]{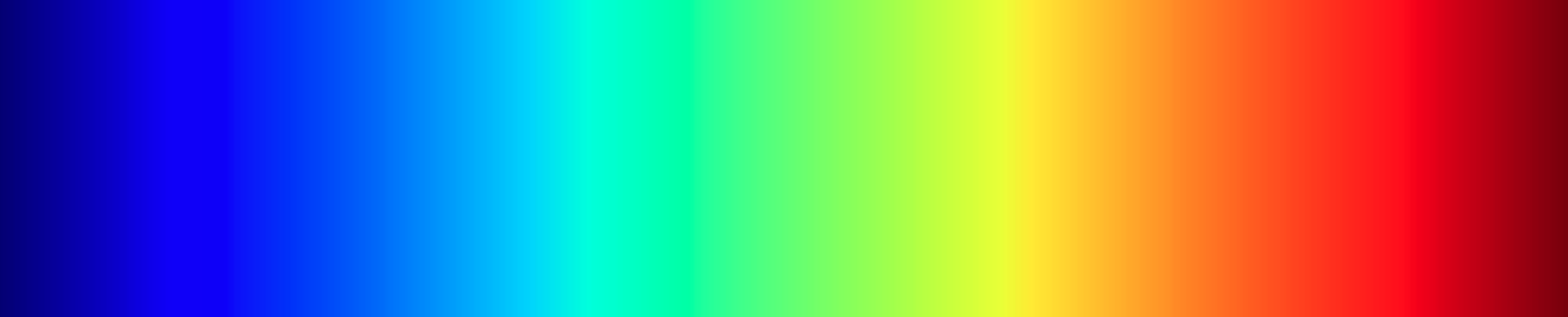} 20cm).}
   \label{fig:cape512-result}
\end{figure*}

Since the ground-truth flow is provided in CAPE-512, we can use it to quantitatively compare the performance of different methods. As shown in Table \ref{tab:cape512-result}, our method achieves state-of-the-art performance in EPE3D, accuracy and outlier ratio. Even without test refinement, it has 28.2\% lower error compared to the nearest baseline. Our method also exhibits a much stronger ability to reduce outliers. Notably, our method obtains a great boost with the help of test refinement, making EPE3D, AccS, AccR and outlier ratio better by 29.1\%, 28.3\%, 1.8\% and 9.8\%, respectively. This proves that our part-rigid assumption is valid for registering human point clouds.

Referring to the visualization results in Fig. \ref{fig:cape512-result}, HumanReg can best align the human point clouds, especially for extremities and large moving parts. This is because our method introduces body part information, which can jointly optimize the points of a certain part. The comparison results on sparse point clouds demonstrate the effectiveness of our ideas, and it also shows that our method can successfully densify point clouds by aligning adjacent frames.

Above results show our method can achieve remarkable performance under the setting of 512 points. Although this can simulate most real-world scenarios, we continue to reduce the number of points to verify the robustness of our method. Specifically, we continue to sample 256 and 128 points from the original CAPE dataset. The results are shown in Table \ref{tab:cape256-128-result}. Our method has comparable performance on sparser point clouds with other baselines on CAPE-512. It's worth noting that ALL other self-supervised baselines failed to converge when the number of points dropped below 512. This proves that our method is more robust to sparse points inputs.

\begin{table}
   \caption{\textbf{Results on sparser point clouds of CAPE dataset.} Besides CAPE-512, we continue to sample 256 and 128 points from the original CAPE dataset and report the registration results.} 
   \begin{center}
   \resizebox{0.75\linewidth}{!}{
   \begin{tabular}{ccccc}
   \toprule
   \# Points & EPE3D & AccS & AccR & Outlier \\
   \midrule
   512       & 3.22  & 85.6 & 97.4 & 0.46    \\
   256       & 6.19  & 47.0 & 88.4 & 1.59    \\
   128       & 8.25  & 28.5 & 73.1 & 3.11    \\
   \bottomrule
   \end{tabular}}
   \end{center}
   \label{tab:cape256-128-result}
\end{table}

\subsection{Qualitative Results on BasketballPlayer}
\begin{figure}
   \centering
   \includegraphics[width=\linewidth]{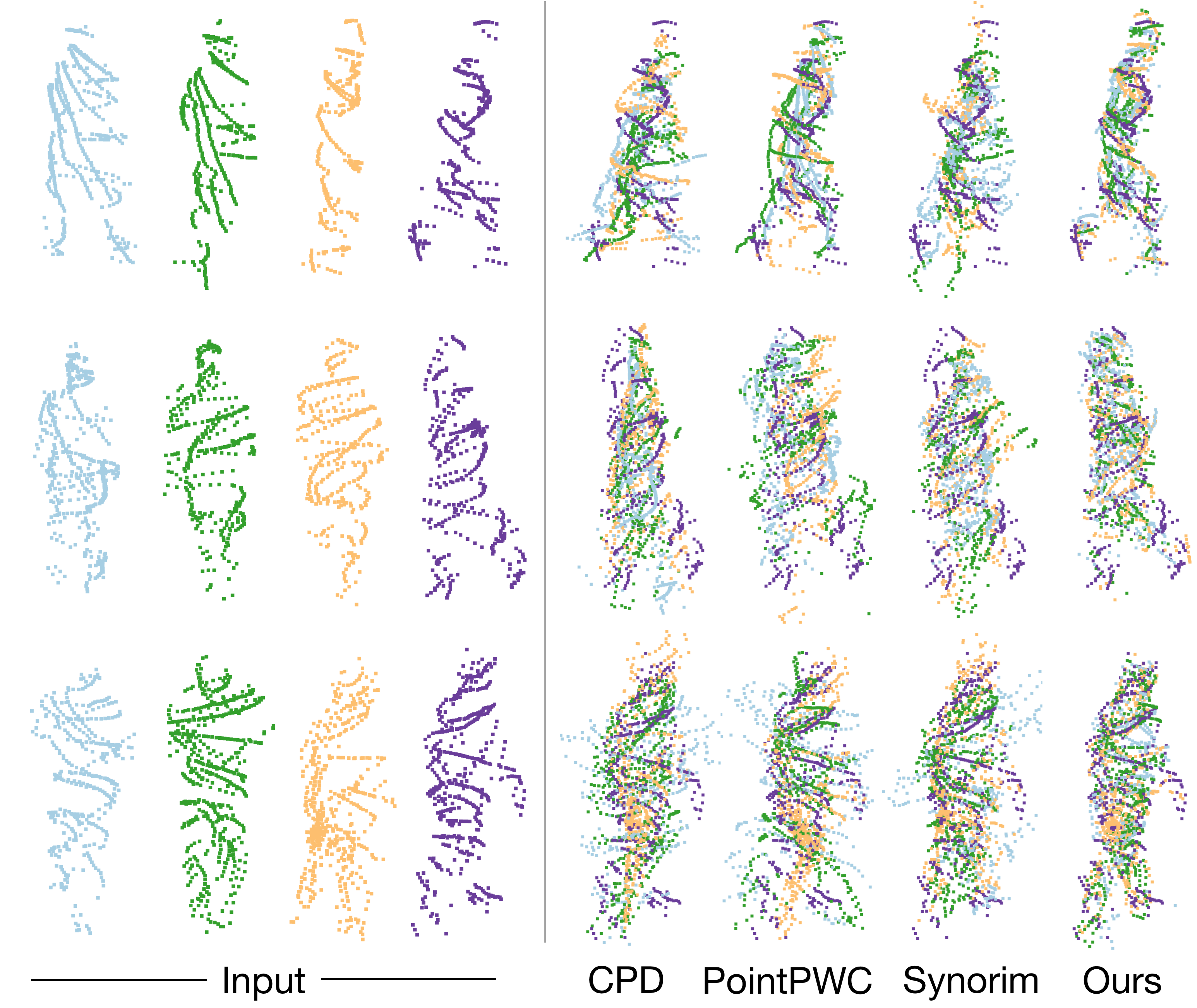}
   \caption{\textbf{Qualitative comparison results on BasketballPlayer.} Different colors are used to distinguish different frames.}
   \label{fig:basketball-result}
\end{figure}

BasketballPlayer is a much more challenging dataset with multiple people moving quickly in a large scene. The performance on it reflects the alignment ability of the human point clouds collected in real outdoor scenes. Due to the lack of ground truth, we only made a qualitative comparison as shown in Fig. \ref{fig:basketball-result}. The results show that our method can more accurately densify the human point clouds and ensure the correctness of human shape.

\begin{table*}
\caption{\textbf{Comparison of different training strategies and learning loss of HumanReg.} We show the results of five sets of experiments with different training strategies and losses on CAPE-512 as ablation studies.}
\begin{center}
\resizebox{0.85\linewidth}{!}{
\begin{tabular}{ccccccccccc}
\toprule
\multirow{2}{*}{No.} & \multirow{2}{*}{\begin{tabular}[c]{@{}c@{}}Pretrain on\\ HumanSyn4D\end{tabular}} & \multicolumn{4}{c}{Self-supervised Learning Loss} & \multirow{2}{*}{\begin{tabular}[c]{@{}c@{}}Test\\ Refinement\end{tabular}} & \multicolumn{4}{c}{Test Metrics} \\ \cmidrule(lr){3-6} \cmidrule(lr){8-11} 
                    &                                                                                   & Chamfer   & Smooth  & Cluster  & Non-rigid  &                                                                            & EPE3D  & AccS  & AccR  & Outlier \\ \midrule
1                    & -                                                                                 & \checkmark         & \checkmark           & -           & -          & -                                                                          & 8.69   & 40.1  & 74.9  & 8.08    \\
2                    & -                                                                                 & \checkmark         & \checkmark           & \checkmark           & \checkmark          & \checkmark                                                                          & 6.91   & 57.4  & 81.5  & 5.50    \\
3                    & \checkmark                                                                                 & -         & -           & -           & -          & \checkmark                                                                          & 7.68   & 61.5  & 78.9  & 10.83   \\
4                    & \checkmark                                                                                 & \checkmark         & \checkmark           & -           & -          & -                                                                          & 5.03   & 58.6  & 94.2  & 0.57    \\
5                    & \checkmark                                                                                 & \checkmark         & \checkmark           & \checkmark           & \checkmark          & \checkmark                                                                          & \textbf{3.22}   & \textbf{85.6}  & \textbf{97.4}  & \textbf{0.46}    \\ \bottomrule
\end{tabular}
}
\end{center}
\label{tab:ablation-studies}
\end{table*}

\section{Ablation Studies}
We conducted five sets of experiments on CAPE-512, with the settings and results are shown in Table \ref{tab:ablation-studies}. Among them, No.1 and No.2 use the data of CAPE-512 to train from scratch  without pretraining process. No.3 uses pretrained model and tests it on CAPE-512 without further finetuning. No.4 and No.5 follow the complete process from pretraining to finetuning. 

\noindent \textbf{Ablation on HumanSyn4D Dataset.} Comparing the result of No.2 and No.5, the pretraining process reduces the EPE3D by 53.4\% and improves the strict accuracy rate from 57.4\% to 85.6\%. This is mainly because learning body-part segmentation directly on unlabeled data is difficult. Since we add a sufficient number of human poses to our synthetic dataset, pretraining can make the model have a good initial value and converge on new datasets.

\noindent \textbf{Ablation on Finetuning Process.} It is not feasible to directly transfer from the synthetic dataset to the real dataset. The No.3 experiment shows a large outlier ratio caused by the difference in data modality between the two datasets. In CAPE-512, the points are randomly sampled from the raw scan, while our synthetic dataset simulates points scanned with mechanical LiDARs. Therefore, a fine-tuning step is necessary for new data.

\noindent \textbf{Ablation on Clustering and Non-rigid Loss.} Comparing the result of No.4 and No.5, our proposed self-supervised loss reduces EPE3D and outlier ratio by 34.0\% and 19.3\%, while improving AccS and AccR by 46.1\% and 3.4\%. This demonstrates the importance of incorporating body part information for registering human point clouds.
\section{Conclusion}
In this work we propose HumanReg, a non-rigid registration method for sparse human point cloud. HumanReg combines flow estimation task with body-part segmentation, which makes correspondence matching based on point features more robust. We also introduce a novel self-supervised loss to our framework and make it possible to learn from unlabeled data. To train our model, we synthesize a labeled dataset, HumanSyn4D, and pretrain HumanReg on it. Then we finetune it on new unlabeled datasets in a self-supervised manner. The experiments show that our framework achieves state-of-the-art performance on CAPE-512 and gains satisfactory results on BasketballPlayer.

\noindent\textbf{Limitations and Future Work.} Despite the state-of-the-art performance, a few limitations are yet to be addressed: (1) Our method is based on pair-wise matching when aligning a sequence of point clouds, which ignores temporal information. An optimization method like \cite{huang2022multiway} or temporal feature extraction module can be added in our framework. (2) Registration task on sparse point cloud is still particularly challenging. Our method still suffers from some failures when faced with real outdoor point clouds due to its sparsity and motion noise. (3) The improvement of our method to the performance of downstream tasks has yet to be proved by further experiments.

{\small
\bibliographystyle{ieeenat_fullname}
\bibliography{egbib}
}

\end{document}